%% file: iclr2022_conference.tex
\documentclass{article} % For LaTeX2e
\usepackage{iclr2022_conference,times}
\usepackage{csquotes}

\input{math_commands.tex}

\usepackage{multirow}
\usepackage{array}
\usepackage{hyperref}
\usepackage{url}
\usepackage{graphicx}

\title{	
Autoregressive Renaissance in Neural PDE Solvers}

\author{Yolanne Yi Ran Lee \\
Department of Computer Science\\
University College London\\
Gower Street, London WC1E 6BT, UK \\
\texttt{yolanne.lee.19@ucl.ac.uk}
}

\iclrfinalcopy % Uncomment for camera-ready version, but NOT for submission.
\begin{document}

\maketitle

\begin{abstract}
Recent developments in the field of neural partial differential equation (PDE) solvers have placed a strong emphasis on neural operators. However, the paper "Message Passing Neural PDE Solver" by Brandstetter et al. published in ICLR 2022 revisits autoregressive models and designs a message passing graph neural network that is comparable with or outperforms both the state-of-the-art Fourier Neural Operator and traditional classical PDE solvers in its generalization capabilities and performance. This blog post delves into the key contributions of this work, exploring the strategies used to address the common problem of instability in autoregressive models and the design choices of the message passing graph neural network architecture.
\footnote{Interactive web version 
 published at \href{https://iclr-blogposts.github.io/2023/blog/2023/autoregressive-neural-pde-solver/}{https://iclr-blogposts.github.io/2023/blog/2023/autoregressive-neural-pde-solver/}.}
\end{abstract}

\section{Introduction}
\renewcommand{\arraystretch}{1.5} % Increase row spacing

\emph{Improving PDE solvers has trickle down benefits to a vast range of other fields.}

Partial differential equations (PDEs) play a crucial role in modeling complex systems and understanding how they change over time and in space.

They are used across physics and engineering, modeling a wide range of physical phenomena like heat transfer, sound waves, electromagnetism, and fluid dynamics, but they can also be used in finance to model the behavior of financial markets, in biology to model the spread of diseases, and in computer vision to model the processing of images.

They are particularly interesting in deep learning!

\begin{enumerate}
    \item Neural networks can be used to model complex PDEs.
    \item Embedding knowledge of a PDE into a neural network can help it generalize better and/or use less data.
    \item PDEs can help explain, interpret, and design neural networks.
\end{enumerate}

Despite their long history dating back to equations first formalized by Euler over 250 years ago, finding numerical solutions to PDEs continues to be a challenging problem.

The recent advances in machine learning and artificial intelligence have opened up new possibilities for solving PDEs in a more efficient and accurate manner. These developments have the potential to revolutionize many fields, leading to a better understanding of complex systems and the ability to make more informed predictions about their behavior.

The background and problem set up precedes a brief look into classical and neural solvers, and finally discusses the message passing neural PDE solver (MP-PDE) introduced by~\citet{brandstetterMessagePassingNeural2022}.

\section{Background}
\subsection{Let's brush up on the basics...}

\emph{The notation and definitions provided match those in the paper for consistency, unless otherwise specified.}

Ordinary differential equations (ODEs) describe how a function changes with respect to a single independent variable and its derivatives. In contrast, PDEs are mathematical equations that describe the behavior of a dependent variable as it changes with respect to multiple independent variables and their derivatives.

Formally, for one time dimension and possibly multiple spatial dimensions denoted by \(\textbf{x}=[x_{1},x_{2},x_{3},\text{...}]^{\top} \in \mathbb{X}\), a general (temporal) PDE may be written as

\begin{equation}
\partial_{t}\textbf{u}= F\left(t, \textbf{x}, \textbf{u},\partial_{\textbf{x}}\textbf{u},\partial_{\textbf{xx}}\textbf{u},\text{...}\right) \qquad (t,\mathbf{x}) \in [0,T] \times \mathbb{X}
\end{equation}

The \(\partial\) is a partial derivative operator which can be understood as "a small change in". For example, the \(\partial_{t}\textbf{u}\) term refers to how much an infinitesmally small change in \(t\) changes \(\textbf{u}\). Below is an explicit definition for some arbitrary function \(f(x,y)\):

\begin{equation}
\frac{\partial f(x,y)}{\partial x} = \lim_{h \to 0} \frac{f(x+h,y) - f(x,y)}{h}
\end{equation}

\begin{itemize}
    \item Initial condition:
 \(\mathbf{u}(0,\mathbf{x})=\mathbf{u}^{0}(\mathbf{x})\) for \(\mathbf{x} \in \mathbb{X}\)
    \item Boundary conditions:
 \(B[ \mathbf{u}](t,x)=0\) for \((t,\mathbf{x}) \in [0,T] \times \partial \mathbb{X}\)
\end{itemize}

\paragraph{Types of boundary conditions} Dirichlet boundary conditions prescribe a fixed value of the solution at a particular point on the boundary of the domain. Neumann boundary conditions, on the other hand, prescribe the rate of change of the solution at a particular point on the boundary. There are also mixed boundary conditions, which involve both Dirichlet and Neumann conditions, and Robin boundary conditions, which involve a linear combination of the solution and its derivatives at the boundary.

The study of PDEs is in itself split into many broad fields. Briefly, these are two other important properties in addition to the initial and boundary conditions:

\paragraph{Linearity}
\begin{itemize}
    \item Linear: the highest power of the unknown function appearing in the equation is one (i.e., a linear combination of the unknown function and its derivatives)
    \item Nonlinear: the highest power of the unknown function appearing in the equation is greater than one
\end{itemize}

\paragraph{Homogeneity}
For an example PDE \(u_t - u_xx = f(x, t)\):
\begin{itemize}
    \item Homogeneous: PDEs with no constant terms (i.e., the right-hand side \(f(x,t)=0\)) and express a balance between different physical quantities
    \item Inhomogeneous: PDEs with a non-zero constant term \(f(x,t)\neq0\) on the right-hand side and describe how an external factor affects the balance
\end{itemize}

PDEs can be either linear or nonlinear, homogeneous or inhomogeneous, and can contain a combination of constant coefficients and variable coefficients. They can also involve a variety of boundary conditions, such as Dirichlet, Neumann, and Robin conditions, and can be solved using analytical, numerical, or semi-analytical methods~\citep{straussPartialDifferentialEquations2007}.

\citet{brandstetterMessagePassingNeural2022} follow precedence set by~\citet{liFourierNeuralOperator2021} and~\citet{bar-sinaiLearningDatadrivenDiscretizations2019} to focus on PDEs written in conservation form:

\begin{equation}
\partial_{t} \mathbf{u} + \nabla \cdot \mathbf{J}(\mathbf{u}) = 0
\end{equation}

\begin{itemize}
    \item \(J\) is the flux, or the amount of some quantity that is flowing through a region at a given time
    \item \(\nabla \cdot J\) is the divergence of the flux, or the amount of outflow of the flux at a given point
\end{itemize}

Additionally, they consider Dirichlet and Neumann boundary conditions.

\subsection{Solving PDEs the classical way}
A brief search in a library will find numerous books detailing how to solve various types of PDEs.

\paragraph{Analytical methods: an exact solution to a PDE can be found by mathematical means~\citep{straussPartialDifferentialEquations2007}}

\begin{itemize}
    \item Separation of Variables
    \begin{itemize}
        \item This method involves expressing the solution as the product of functions of each variable, and then solving each function individually. It is mainly used for linear PDEs that can be separated into two or more ordinary differential equations.
    \end{itemize}
    \item Green's Functions
    \begin{itemize}
        \item This method involves expressing the solution in terms of a Green's function, which is a particular solution to a homogeneous equation with specified boundary conditions.
    \end{itemize}
\end{itemize}

\paragraph{Semi-analytical methods: an analytical solution is combined with numerical approximations to find a solution~\citep{bartelsNumericalApproximationPartial}}

\begin{itemize}
    \item Perturbation methods
    \begin{itemize}
        \item This method is used when the solution to a PDE is close to a known solution or is a small deviation from a known solution. The solution is found by making a perturbation to the known solution and solving the resulting equation analytically.
    \end{itemize}
    \item Asymptotic methods
    \begin{itemize}
        \item In this method, the solution is represented as a series of terms that are solved analytically. The solution is then approximated by taking the leading terms of the series.
    \end{itemize}
\end{itemize}

\emph{Very few PDEs have analytical solutions, so numerical methods have been developed to approximate PDE solutions over a wider range of potential problems.}

\subsubsection{Numerical Methods}
Often, approaches for temporal PDEs follow the method of lines (MOL).

Every point of the discretization is then thought of as a separate ODE evolving in time, enabling the use of ODE solvers such as Runge-Kutta methods.

\paragraph{1. Discretizing the problem}

In the most basic case (a regular grid), arbitrary spatial and temporal resolutions \(\mathbf{n_{x}}\) and \(n_{t}\) can be chosen and thus used to create a grid where \(\mathbf{n_{x}}\) is a vector containing a resolution for each spatial dimension.

The domain may also be irregularly sampled, resulting in a grid-free discretization. This is often the case with real-world data that comes from scattered sensors, for example.

Finite difference methods (FDMs) or any other discretization technique can be used to discretize the time domain.

One direction of ongoing research seeks to determine discretization methods which can result in more efficient numerical solvers (for example, take larger steps in flatter regions and smaller steps in rapidly changing regions).

\paragraph{2. Estimating the spatial derivatives}
A popular choice when using a gridded discretization is the finite difference method (FDM). Spatial derivative operators are replaced by a stencil which indicates how values at a finite set of neighboring grid points are combined to approximate the derivative at a given position. This stencil is based on the Taylor series expansion.

The finite volume method (FVM) is another approach which works for irregular geometries. Rather than requiring a grid, the computation domain can be divided into discrete, non-overlapping control volumes used to compute the solution for that portion~\citep{bartelsNumericalApproximationPartial}.

For every control volume, a set of equations describing the balance of some physical quantities (in essence, estimating the flux at control volume boundaries) can be solved which results in the approximated spatial derivative.

While this method only works for conservation form equations, it can handle complex problems with irregular geometries and fluxes that are difficult to handle with other numerical techniques such as the FDM.

In the pseudospectral method (PSM), PDEs are solved pointwise in physical space by using basis functions to approximate the spatial derivatives. The pseudospectral method and the Galerkin method are two common examples of spectral methods which use basis functions satisfying various conditions depending on the specific algorithm being applied. While the FDM considers local information to construct approximations, spectral methods determine global solutions and have exponential convergence.

These methods are well-suited for solving problems with smooth solutions and periodic boundary conditions, but their performance drops for irregular or non-smooth solutions, as well as problems with more degrees of freedom where their global nature results in high dimensional dense matrix computations.

\paragraph{3. Time updates}
The resulting problem is a set of temporal ODEs which can be solved with classical ODE solvers such as any member of the Runge-Kutta method family.

\subsubsection{Limitations of Classical Methods}
The properties of a PDE, such as its order, linearity, homogeneity, and boundary conditions, determine its solution method. Different methods have been developed based on the different properties and requirements of the problem at hand.~\citet{brandstetterMessagePassingNeural2022} categorizes these requirements into the following:

\begin{table}
\centering
\begin{tabular}{|p{0.3\linewidth}|p{0.3\linewidth}|p{0.3\linewidth}|}
\hline
\textbf{User} & \textbf{Structural} & \textbf{Implementational} \\
\hline
Computation efficiency, computational cost, accuracy, guarantees (or uncertainty estimates), generalization across PDEs & Spatial and temporal resolution, boundary conditions, domain sampling regularity, dimensionality & Stability over long rollouts, preservation of invariants \\
\hline
\end{tabular}
% \caption{Your caption here} % Optional
\end{table}

The countless combinations of requirements resulted in what Bartels defines as a splitter field~\citep{bartelsNumericalApproximationPartial}: a specialized classical solver is developed for each sub-problems, resulting in many specialized tools rather than a single one.

These methods, while effective and mathematically proven, often come at high computation costs. Taking into account that PDEs often exhibit chaotic behaviour and are sensitive to any changes in their parameters, re-running a solver every time a coefficient or boundary condition changes in a single PDE can be computationally expensive.

One key example which limits grid-based classical solvers is the Courant-Friedrichs-Lewy (CFL) condition, which states that the maximum time step size should be proportional to the minimum spatial grid size. According to this condition, as the number of dimensions increases, the size of the temporal step must decrease and therefore numerical solvers become very slow for complex PDEs.

\begin{table}[htb]
\centering
\begin{tabular}{|p{0.4\linewidth}|p{0.1\linewidth}|p{0.1\linewidth}|l|}
\hline
\textbf{Algorithm} & \textbf{Equation} & \textbf{Boundary conditions} & \textbf{Complexity} \\
\hline
Classical FDM/FEM/FVM & general & general & \( \text{poly} \left( \left( \frac{1}{\varepsilon} \right)^{d} \right) \) \\
\hline
Adaptive FDM/FEM~\citep{babuskaHpVersionFinite1987} & general & general & \( \text{poly} \left( \left( \log \left( \frac{1}{\varepsilon} \right) \right)^{d} \right) \) \\
\hline
Spectral method~\citep{gheorghiuSpectralMethodsDifferential2007,shenSpectralMethodsAlgorithms2011} & general & general & \( \text{poly} \left( \left( \log \left( \frac{1}{\varepsilon} \right) \right)^{d} \right) \) \\
\hline
Sparse grid FDM/FEM~\citep{bungartzSparseGrids2004,zengerSparseGrids1991} & general & general & \( \text{poly} \left( \frac{1}{\varepsilon} \left( \log \left( \frac{1}{\varepsilon} \right) \right)^{d} \right) \) \\
\hline
Sparse grid spectral method~\citep{shenEfficientSpectralSparse2010} & elliptic & general & \( \text{poly} \left( \log \left( \frac{1}{\varepsilon} \right) \left( \log \log \left( \frac{1}{\varepsilon} \right) \right)^{d} \right) \) \\
\hline
\end{tabular}
\caption{Table showing (polynomial) computational complexity of some common numerical methods, including finite difference method (FDM), finite elements method (FEM), finite volume method (FVM), spectral method, and some of their variants for \(d\)-dimensional PDEs with error tolerance \(\epsilon\). Note that every method has an exponential dependency on the dimensionality. Adapted from~\citep{childsHighprecisionQuantumAlgorithms2021}} % Optional
\end{table}

\subsection{Neural Solvers}

Neural solvers offer some very desirable properties that may serve to unify some of this splitter field. Neural networks can learn and generalize to new contexts such as different initial/boundary conditions, coefficients, or even different PDEs entirely~\citep{brandstetterMessagePassingNeural2022}. They can also circumvent the CFL condition, making them a promising avenue for solving highly complex PDEs such as those found in weather prediction. For a review which contextualizes physics informed machine learning with regards to classical problems and methods, see~\citep{mengWhenPhysicsMeets2022}.

Though most methods lie along a spectrum from classical leaning to end-to-end neural, a naive yet illustrative categorization into three groupings is shown below. 

\begin{figure}[htb]
    \centering
    \includegraphics[width=\linewidth]{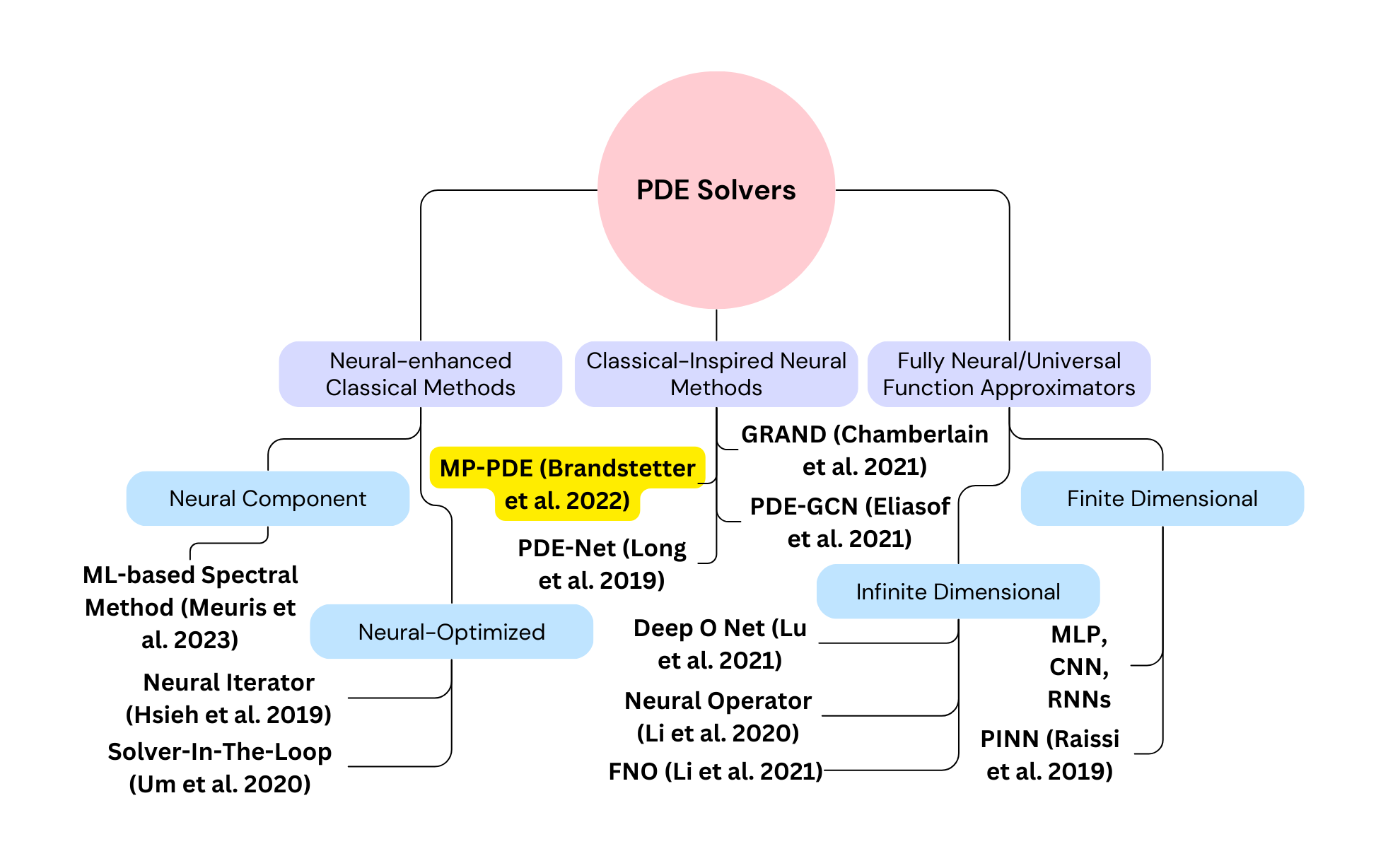}
    % \caption{Caption}
    \label{fig:PDEchart}
\end{figure}

\subsubsection{Fully Neural/Universal Function Approximators}

The term fully neural here refers to methods which rely on the universal function approximation theory such that a sufficiently complex network can represent any arbitrary function. Many common fully neural methods are also known as neural operators which model the solution of a PDE as an operator that maps inputs to outputs. The problem is set such that a neural operator $\mathcal{M}$ satisfies $\mathcal{M}(t,\mathbf{u}^{0}) = \mathbf{u}(t)$ where $\mathbf{u}^{0}$ are the initial conditions~\citep{luDeepONetLearningNonlinear2021, brandstetterMessagePassingNeural2022}. The idea of using deep learning techniques to solve differential equations has a long history, including Dissanayake's and Phan-Thien's attempt to use multilayer perceptrons (MLPs) as universal approximators to solve PDEs, and arguably includes any work involving incorporating prior knowledge into models in general~\citep{dissanayakeNeuralnetworkbasedApproximationsSolving1994,psichogiosHybridNeuralNetworkfirst1992,lagarisArtificialNeuralNetworks1998}. Simple MLPs, CNNs, RNNs, and other networks used to map input vectors to output vectors are naive examples of finite-dimensional operators.

\citet{raissiPhysicsinformedNeuralNetworks2019} officially coined the physics-informed neural network (PINN) in 2017. The problem is set such that the network $$\mathcal{N}$$ satisfies $$\mathcal{N}(t,\mathbf{u}^{0}) = \mathbf{u}(t)$$ where $$\mathbf{u}^{0}$$ are the initial conditions. The main principle behind PINNs is to enforce the governing physical laws of the problem on the network's predictions by adding loss term(s) to the network's objective function.

For parameters $\theta = \text{argmin}_{\theta} \mathcal{L}(\theta)$ where $\mathcal{L}$ is a loss function and $\lambda$ is some weighting, the loss with a physics prior may be defined as follows, where the terms are described in \ref{table:pinnterms}:

the loss with a physics prior may be defined as follows:

\begin{equation}
\mathcal{L}(\theta) = \omega_{\mathcal{F}} \mathcal{L}_{\mathcal{F}}(\theta) + \omega_{\mathcal{B}} \mathcal{L}_{\mathcal{B}}(\theta) + \omega_{d} \mathcal{L}_{\text{data}}(\theta)
\end{equation}

\begin{table}[ht]
\centering
\begin{tabular}{|p{2.5cm}|p{4cm}|p{5cm}|}
\hline
Term & Definition & Effect \\
\hline
$\mathcal{L}_{\mathcal{B}}$ & Loss wrt. the initial and/or boundary conditions & Fits the known data over the network \\
\hline
$\mathcal{L}_{\mathcal{F}}$ & Loss wrt. the PDE & Enforces DE $\mathcal{F}$ at collocation points; Calculating using autodiff to compute derivatives of $\mathbf{\hat{u}_{\theta}(\mathbf{z})}$ \\
\hline
$\mathcal{L}_{\text{data}}$ & Validation of known data points & Fits the known data over the NN and forces $\mathbf{\hat{u}}_{\theta}$ to match measurements of $\mathbf{u}$ over provided points \\
\hline
\end{tabular}
\caption{PINN formulation terms and descriptions.}
\label{table:pinnterms}
\end{table}

Since the network maps input variables to output variables which are both finite-dimensional and dependent on the grid used to discretize the problem domain, it is considered a finite dimensional neural operator. The paper gained a lot of traction and inspired many architectures which now fall under the PINN family; for a more thorough review see~\citep{cuomoScientificMachineLearning2022}, and for hands-on examples visit the digital book by~\citep{thuereyPhysicsbasedDeepLearning2022}.

The success of this loss-based approach is apparent when considering the rapid growth of papers which extend the original iteration of the PINN. However,~\citet{krishnapriyanCharacterizingPossibleFailure2021} has shown that even though standard fully-connected neural networks are theoretically capable of representing any function given enough neurons and layers, a PINN may still fail to approximate a solution due to the complex loss landscapes arising from soft PDE constraints.

The DeepONet architecture is a seminal example of an infinite dimensional neural operator in contrast to the finite dimensional PINN~\citep{luDeepONetLearningNonlinear2021}. It consists of one or multiple branch net(s) which encode discrete inputs to an input function space, and a single trunk net which receives the query location to evaluate the output function. The model maps from a fixed, finite dimensional grid to an infinite dimensional output space.

Since the development of the DeepONet, many novel neural operators have emerged which generalize this finite-infinite dimensional mapping to an infinite-infinite dimensional mapping~\citep{liNeuralOperatorGraph2020,liPhysicsinformedNeuralOperator2021,goswamiPhysicsInformedDeepNeural2022,rahmanUshapedNeuralOperators2022,tripuraWaveletNeuralOperator2022,fanaskovSpectralNeuralOperators2022,pathakFourCastNetGlobalDatadriven2022}, including the Fourier Neural Operator (FNO). It operates within Fourier space and takes advantage of the convolution theorem to place the integral kernel in Fourier space as a convolutional operator.

These global integral operators (implemented as Fourier space convolutional operators) are combined with local nonlinear activation functions, resulting in an architecture which is highly expressive yet computationally efficient, as well as being resolution-invariant.

While the vanilla FNO required the input function to be defined on a grid due to its reliance on the FFT, further work developed mesh-independent variations as well~\citep{liPhysicsinformedNeuralOperator2021}.

\begin{figure}
    \centering
    \includegraphics[width=\linewidth]{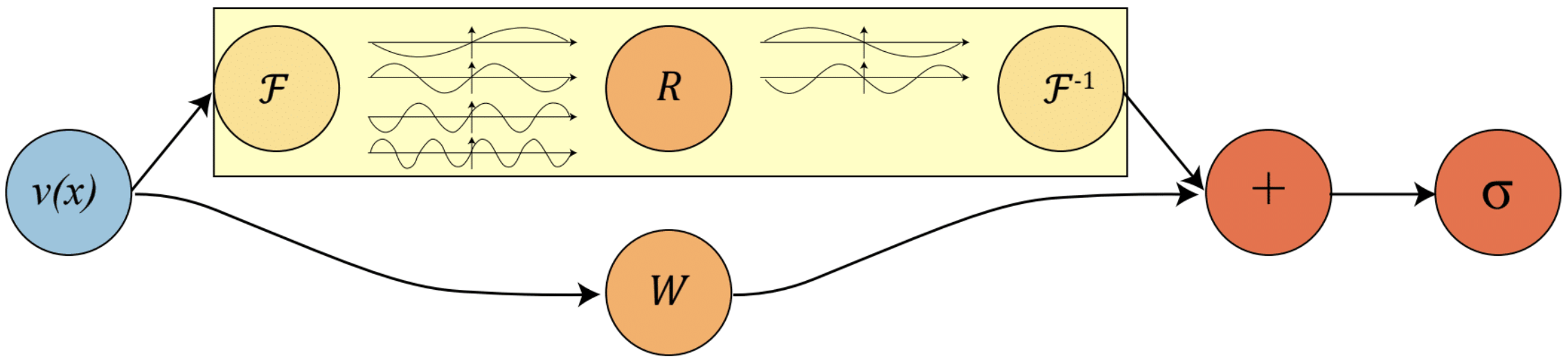}
    \caption{FNO architecture. For more details, see \href{https://zongyi-li.github.io/blog/2020/fourier-pde/}{this blogpost}. Credits:~\citet{liFourierNeuralOperator2021}}
    \label{fig:FNO}
\end{figure}

Neural operators are able to operate on multiple domains and can be completely data-driven.

However, these models do not tend to predict out-of-distribution \(t\) and are therefore limited when dealing with temporal PDEs. Another major barrier is their relative lack of interpretability and guarantees compared to classical solvers. 

\subsection{Neural-Augmented Classical Methods}

A parallel line of research involves using deep learning as a tool to improve classical numerical methods for solving PDEs. One avenue involves modifying existing iterative methods: while neural operator methods directly mapped inputs to outputs, autoregressive methods take an iterative approach instead. For example, iterating over time results in a problem such as
\[
\mathbf{u}(t+\Delta t) = \mathcal{A}(\Delta t, \mathbf{u}(t))
\]
where $\mathcal{A}$ is some temporal update~\citep{brandstetterMessagePassingNeural2022}.

\begin{figure}[ht]
\centering
\includegraphics[width=0.45\textwidth]{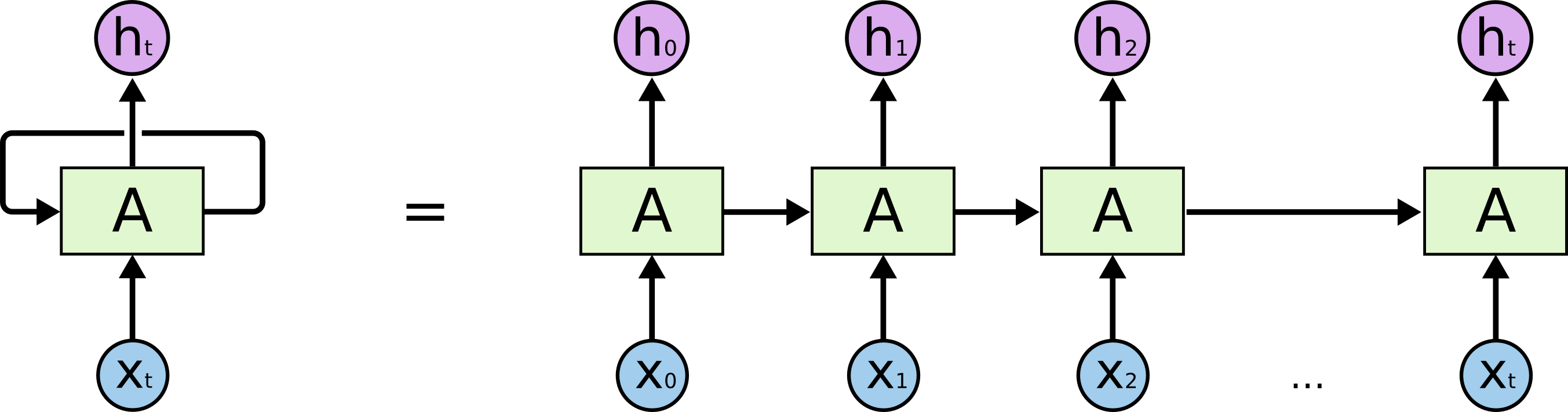}
\includegraphics[width=0.45\textwidth]{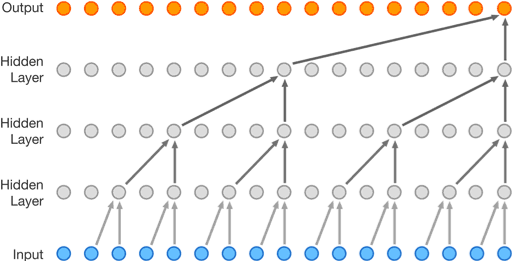}
\caption{Similarly to RNNs (left), autoregressive models take previous time steps to predict the next time step. However, autoregressive models (right) are entirely feed-forward and take the previous predictions as inputs rather than storing them in some hidden state. Credits: RNN diagram from Colah's Blog~\citep{UnderstandingLSTMNetworks}, WaveNet from Deepmind Blog~\citep{WaveNetGenerativeModel}.}
\end{figure}

Popular autoregressive models include PixelCNN for images, WaveNet for audio, and the Transformer for text.

Three autoregressive systems mentioned by Brandstetter et al.\ are hybrid methods which use neural networks to predict certain parameters for finite volume, multigrid, and iterative finite elements methods. All three retain a (classical) computation grid which makes them somewhat interpretable~\citep{bar-sinaiLearningDatadrivenDiscretizations2019, greenfeldLearningOptimizeMultigrid2019a, hsiehLearningNeuralPDE2019}.

\citet{hsiehLearningNeuralPDE2019}, for example, develops a neural network-accelerated iterative finite elements method. Most significantly, their approach offers theoretical guarantees of convergence and correctness. Their problem formulation focuses on solving a single linear PDE class for variable discretization, boundary conditions, and source/forcing terms. For any PDE with an existing linear iterative solver, a learned iterator can replace a handcrafted classical iterator.

Similarly,~\citet{umSolverintheLoopLearningDifferentiable2020} proposed using a neural network component to learn the error or deviation from the path of an iterative solver. Using this component, the iterative method can be "pushed" back onto the true PDE solution.

Another way deep learning can be leveraged in classical methods is characterized by~\citep{meurisMachinelearningbasedSpectralMethods2023} and also highlights the deeply interconnected nature of these novel developments. The conventional spectral method rewrites a PDE in terms of the sum of basis functions; Meuris et al.\ use a DeepONet to discover candidate functions to be used as basis functions. Though mathematical work is required to mold the extracted function (from the DeepONet) to a basis function satisfying certain desirable properties, it expands the use of the spectral method toward complex domains where we might not have known appropriate basis functions.

However, augmented classical systems have not gained the acclaim seen by their fully neural counterparts as a whole.

This is on one hand due to their limitations in generalization. In Hsieh et al.'s case, an existing numerical method must be used to craft a complementary neural iterator~\citep{hsiehLearningNeuralPDE2019}. Another major concern is the accumulation of error in iterative methods, which is particularly detrimental for PDE problems that often exhibit chaotic behavior~\citep{brandstetterMessagePassingNeural2022}. Overarching both neural component and neural-optimized methods, however, is the tradeoff between marginal improvements to classical methods and what tends to be a non-trivial amount of manual work required to implement such methods.

\subsection{Classical-Inspired Neural Methods}

Ruthotto and Haber released an impactful study in 2018 which interprets residual neural networks (ResNets) as PDEs, and addresses some of their challenges using PDE theory~\citep{ruthottoDeepNeuralNetworks2018}. A standard ResNet has skip connections which in effect add a previous layer's output directly to the calculation of a future layer's output. Given input features \( \mathbf{Y}_{0} = \mathbf{Y} \) and a ResNet with \( N \) layers, the output of the \( j \)th layer is used to calculate that of the next:

\[ \mathbf{Y}_{j+1} = \mathbf{Y}_{j} + f(\theta^{(j)}, \mathbf{Y}_{j}) \]

This formulation also describes a typical forward Euler discretization with a step size \( \delta_{t} = 1 \). Based on this continuous interpretation of a ResNet layer, PDEs from control theory can be used to develop novel networks with specific and expected behaviours like smoothing or even memory reduction~\citep{ruthottoDeepNeuralNetworks2018}.

This is an example of a strong classical-inspired neural method which allowed us to systematically develop novel architectures. Since then, PDE interpretations of neural network architectures have been expanded to encompass embedding PDEs into architectures themselves, and building architectures to mimic classical PDE solvers.

The Graph Neural Diffusion (GRAND) model introduced by Chamberlain et al.\ demonstrates that graph neural networks (GNNs) can be crafted using differential equations (like diffusion processes) where the spatial derivative is analogous to the difference between node features, and the temporal update is a continuous counterpart to the layer index~\citep{chamberlainGRANDGraphNeural2021a}. From these two principles and their derivations of diffusion PDEs on graphs, Chamberlain et al.\ design networks which ameliorate common GNN pitfalls like oversmoothing (which occurs as the number of layers increases). Note that here, the emphasis is not in outputting the solution of a PDE directly but rather using a PDE to influence or bias the output toward an expected result, somewhat more similarly to how a PINN biases the output to obey a specified PDE.

Later, the PDE-GCN model extends GRAND by deriving differential operators on manifolds which are then discretized on graphs to then build not only diffusion, but hyperbolic PDE-inspired GNNs as well~\citep{eliasofPdegcnNovelArchitectures2021}. The discretized nonlinear diffusion and nonlinear hyperbolic PDEs call back to Ruthotto et al.'s comparison to ResNet updates and are used to define the titular PDE-inspired graph convolutional network (GCN) layer. Interestingly, mixing both diffusion and hyperbolic variants can allow one to discover which is more prominent to a task by retrieving a parameter which weights how much one network dynamic contributes to the output.

This category of models highlights the diverse ways that PDEs are used in deep learning. Not only can these networks be tested on mathematical datasets, but they provide valuable interpretations and performance improvements when used in non-geometric tasks like node classification and even protein-protein interactions in biology.

\section{Message Passing Neural PDE Solver (MP-PDE)}

Brandstetter et al.\ propose a fully neural PDE solver which capitalizes on neural message passing. The overall architecture is laid out below, consisting of an MLP encoder, a GNN processor, and a CNN decoder.

\begin{figure}[htb]
    \centering
    \includegraphics[width=\textwidth]{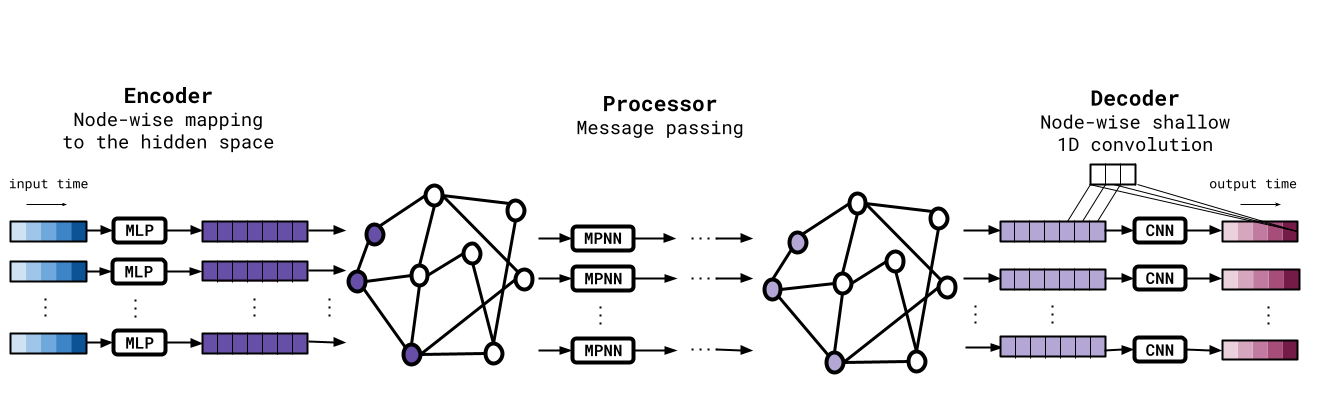}
    \caption{Overall MP-PDE architecture. Credits:~\citet{brandstetterMessagePassingNeural2022}.}
\end{figure}

At its core, this model is autoregressive and thus faces the same challenge listed above. Two key contributions of this work are the pushforward trick and temporal bundling which mitigate the potential butterfly effect of error accumulation~\citep{brandstetterMessagePassingNeural2022}. The network itself, being fully neural, is capable of generalization across many changes as well.

\subsection{The Pushforward Trick and Temporal Bundling}

\subsubsection{Pushforward Trick}
\begin{figure}[htb]
    \centering
    \includegraphics[width=\textwidth]{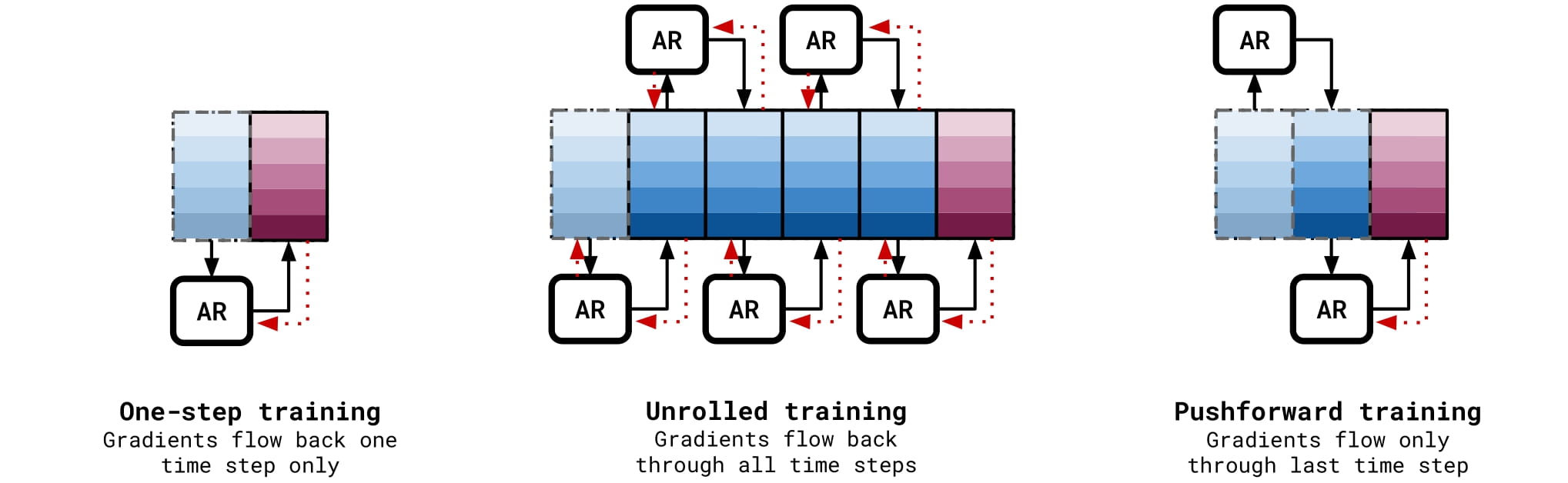}
    \caption{Pushforward trick compared to one-step and unrolled training. Credits: Brandstetter et al.~\citep{brandstetterMessagePassingNeural2022}.}
\end{figure}

During testing, the model uses current time steps (first from data, then from its own predictions) to approximate the next time step.

This results in a distribution shift problem because the inputs are no longer solely from ground truth data: the distribution learned during training will always be an approximation of the true data distribution. The model will appear to overfit to the one-step training distribution and perform poorly the further it continues to predict.

An adversarial-style stability loss is added to the one-step loss so that the training distribution is brought closer to the test time distribution~\citep{brandstetterMessagePassingNeural2022}:

\begin{equation}
L_{\text{one-step}} = \mathbb{E}_{k} \mathbb{E}_{{\mathbf{u^{k+1}}|\mathbf{u^{k}},\mathbf{u^{k} \sim p_{k}}}} \left[ \mathcal{L} \left( \mathcal{A}(\mathbf{u}^{k}), \mathbf{u}^{k+1} \right) \right]
\end{equation}

The loss function is used to evaluate the difference between the temporal update and the expected next state, and the overall one-step loss is calculated as the expected value of this loss over all time-steps and all possible next states.

\begin{equation}
L_{\text{stability}} = \mathbb{E}_{k}\mathbb{E}_{\mathbf{u^{k+1}|\mathbf{u^{k},\mathbf{u^{k} \sim p_{k}}}}}[\mathbb{E}_{\epsilon | \mathbf{u}^{k}} [\mathcal{L}(\mathcal{A}(\mathbf{u}^{k}+\epsilon)),\mathbf{u}^{k+1}]]
\end{equation}

\begin{equation}
L_{\text{total}} = L_{\text{one-step}} + L_{\text{stability}}
\end{equation}

The stability loss is largely based off the one-step loss, but now assumes that the temporal update uses noisy data.

The pushforward trick lies in the choice of \(\epsilon\) such that \(\mathbf{u}^{k}+\epsilon = \mathcal{A}(\mathbf{u}^{k-1})\), similar to the test time distribution. Practically, it is implemented to be noise from the network itself so that as the network improves, the loss decreases.

Necessarily, the noise of the network must be known or calculated to implement this loss term. So, the model is unrolled for 2 steps but only backpropagated over the most recent unroll step, which already has the neural network noise~\citep{brandstetterMessagePassingNeural2022}. In essence, the one-step training has a clean input and noisy output whereas the pushforward trick has both noisy input and noisy output with the \(\epsilon\) term capturing the noise.

While the network could be unrolled during training, this not only slows the training down but also might result in the network learning shortcuts across unrolled steps.

\subsubsection{Temporal Bundling}

\begin{figure}
    \centering
    \begin{minipage}{.65\linewidth}
        \includegraphics[width=\linewidth]{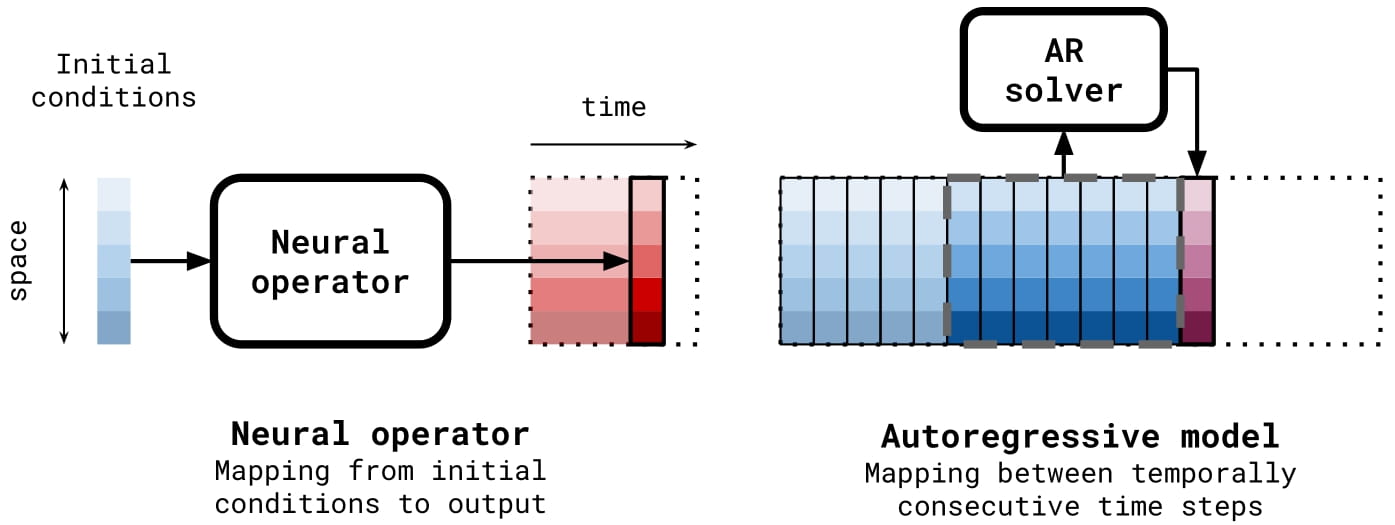}
    \end{minipage}%
    \begin{minipage}{.35\linewidth}
        \includegraphics[width=\linewidth]{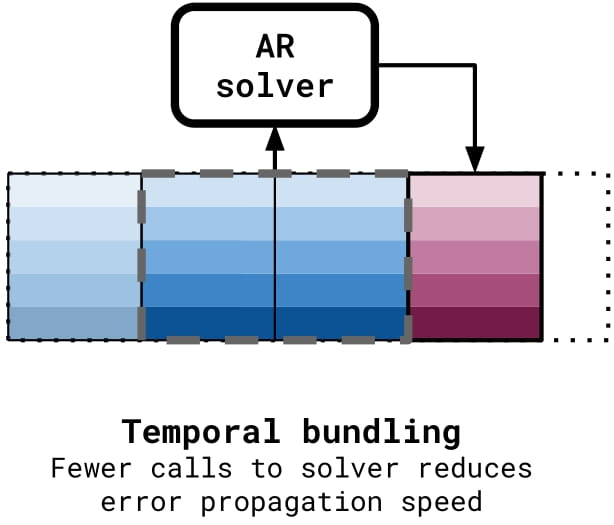}
    \end{minipage}
    \caption{Temporal bundling compared to neural operators and autoregressive models. Credits:~\citet{brandstetterMessagePassingNeural2022}.}
\end{figure}

This trick complements the previous by reducing the amount of times the test time distribution changes. Rather than predicting a single value at a time, the MP-PDE predicts multiple time-steps at a time, as seen above~\citep{brandstetterMessagePassingNeural2022}.

\subsection{Network Architecture}

GNNs have been used as PDE solvers in a variety of works~\citep{liNeuralOperatorGraph2020, eliasofPdegcnNovelArchitectures2021, iakovlevLearningContinuoustimePDEs2021}; however, in this implementation, links can be drawn directly from the MOL to each component of the network architecture centering around the use of a message passing algorithm.

\begin{table}
    \centering
    \begin{tabular}{|l|p{8cm}|}
        \hline
        \textbf{Classical Numerical Method} & \textbf{MP-PDE Network Component} \\
        \hline
        Partitioning the problem onto a grid & Encoder \\
        & \emph{Encodes a vector of solutions into node embeddings} \\
        \hline
        Estimating the spatial derivatives & Processor \\
        & \emph{Estimates spatial derivatives via message passing} \\
        \hline
        Time updates & Decoder \\
        & \emph{Combines some representation of spatial derivatives smoothed into a time update} \\
        \hline
    \end{tabular}
    \caption{Comparison between Classical Numerical Methods and MP-PDE Network Components.}
\end{table}

\paragraph{Encoder}
The encoder is implemented as a two-layer MLP which computes an embedding for each node \(i\) to cast the data to a non-regular integration grid,

\begin{equation}
\mathbf{f}_{i}^{0} = \epsilon^{v}([\mathbf{u}_{i}^{k-K:k},\mathbf{x}_{i},t_{k},\theta_{PDE}])
\end{equation}

where \(\mathbf{u}_{i}^{k-K:k}\) is a vector of previous solutions (the length equaling the temporal bundle length), \(\mathbf{x}_{i}\) is the node's position, \(t_{k}\) is the current timestep, and \(\theta_{PDE}\) holds equation parameters.

\paragraph{Processor}
The node embeddings from the encoder are then used in a message passing GNN. The message passing algorithm, which approximates spatial derivatives, is run \(M\) steps using the following updates:

\begin{center}
\rule{0.6\textwidth}{0.5pt}
\end{center}

\begin{equation}
\text{edge } j \to i \text{ message:} \qquad \mathbf{m}_{ij}^{m} = \phi (\mathbf{f}_{i}^{m}, \mathbf{f}_{j}^{m}, \mathbf{u}_{i}^{k-K:k}-\mathbf{u}_{j}^{k-K:k}, \mathbf{x}_{i}-\mathbf{x}_{j}, \theta_{PDE})
\end{equation}

The difference in spatial coordinates helps enforce translational symmetry and, combined with the difference in node solutions, relates the message passing to a local difference operator. The addition of the PDE parameters is motivated by considering what the MP-PDE should generalize over: by adding this information in multiple places, flexibility can potentially be learned since all this information (as well as the node embeddings) is fed through a two-layer MLP.

In addition, the solution of a PDE at any timestep must respect the boundary condition (the same as in classical methods for BVPs), so adding the PDE parameters in the edge update provides knowledge of the boundary conditions to the neural solver.

\begin{center}
\rule{0.6\textwidth}{0.5pt}
\end{center}

\begin{equation}
\text{node } i \text{ update:} \qquad \mathbf{f}_{i}^{m+1} = \psi (\mathbf{f}^{m}_{i}, \sum_{j \in \mathcal{N}(i)} \mathbf{m}_{ij}^{m}, \theta_{PDE})
\end{equation}

The future node embedding is updated using the current node embedding, the aggregation of all received messages, and (again) the PDE parameters. This information is also fed through a two-layer MLP.

\begin{center}
\rule{0.6\textwidth}{0.5pt}
\end{center}

\citet{bar-sinaiLearningDatadrivenDiscretizations2019} explores the relationship between FDM and FVM as used in the method of lines. In both methods, the \(n^{th}\) order derivative at a point \(x\) is approximated by

\begin{equation}
\partial^{(n)}_{x}u \approx \sum_{i} a^{(n)}_{i} u_{i}
\end{equation},

for some precomputed coefficients \(a^{(n)}_{i}\). The right hand side parallels the message passing scheme, which aggregates the local difference (\(\mathbf{u}_{i}^{k-K:k}-\mathbf{u}_{j}^{k-K:k}\) in the edge update) and other (learned) embeddings over neighborhoods of nodes.

This relationship gives an intuitive understanding of the message passing GNN, which mimics FDM for a single layer, FVM for two layers, and WENO5 (Weighted Essentially Non-Oscillatory (5th order)) for three layers~\citep{brandstetterMessagePassingNeural2022}. WENO5 is a numerical interpolation scheme used to reconstruct the solution at cell interfaces in FVM.

While the interpretation is desirable, how far this holds in the actual function of the MP-GNN (message passing graph neural network) is harder to address. The concepts of the nodes as integration points and messages as local differences break down as the nodes and edges update. In addition, the furthest node that contributes a message from for any point is at \(n\) edges away for the \(n^{th}\) layer (or a specified limit). This results in a very coarse and potentially underinformed approximation for the first layer which is then propagated to the next layers. However, both the updates use two layer MLPs which (although abstracting away from their respective interpretations) may in effect learn optimal weightings to counterbalance this.

\paragraph{Decoder}
The approximated spatial derivatives are then combined and smoothed using a 1D CNN which outputs a bundle of next time steps (recall temporal bundling) \(\mathbf{d}_{i}\). The solution is then updated:

\begin{equation} 
\mathbf{u}^{k+l}_{i} = u^{k}_{i} + (t_{k+l}-t_{k})\mathbf{d}^{l}_{i}
\end{equation}

Some precedence is seen, for example, in classical linear multistep methods which (though effective) face stability concerns. Since the CNN is adaptive, it appears that it avoids this issue~\citep{brandstetterMessagePassingNeural2022}.

\subsection{Results}

\emph{As a general neural PDE solver, the MP-GNN surpasses even the current state-of-the-art FNO.}

For example, after training a neural model and setting up an instance of MOL, this is a brief comparison of how they can generalize without re-training.

\paragraph{Quantitative measures}
\begin{itemize}
    \item Accumulated error: \(\frac{1}{n_{x}} \sum_{x,t} MSE\)
    \item Runtime (s): Measured time taken to run for a given number of steps.
\end{itemize}

\begin{table}
    \centering
    \begin{tabular}{|l|l|l|l|}
        \hline
        \textbf{Generalization to...} & \textbf{MP-GNN} & \textbf{FNO} & \textbf{Classical (MOL)} \\
        \hline
        New PDEs & Yes & No & No \\
        \hline
        Different resolutions & Yes & Yes & No (unless downsampling) \\
        \hline
        Changes in PDE parameters & Yes & Yes & Sometimes \\
        \hline
        Non-regular grids & Yes & Some & Yes (dependent on implementation) \\
        \hline
        Higher dimensions & Yes & No & No \\
        \hline
    \end{tabular}
    \caption{Comparison of generalization capabilities.}
\end{table}

\begin{figure}[htb]
    \centering
    \includegraphics[width=\textwidth]{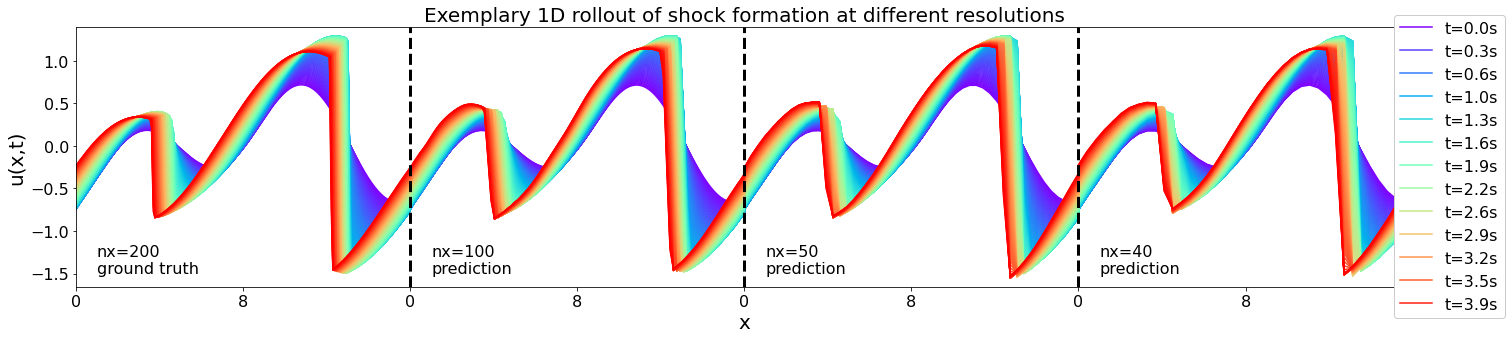}
    \caption{Demonstration of shock formation using MP-PDE from different training data resolutions. Credits:~\citet{brandstetterMessagePassingNeural2022}.}
\end{figure}

This experiment exemplifies the MP-PDE's ability to model shocks (where both the FDM and PSM methods fail) across multiple resolutions. Even at a fifth of the resolution of the ground truth, both the small and large shocks are captured well.

\begin{figure}[htb]
    \centering
    \includegraphics[width=\textwidth]{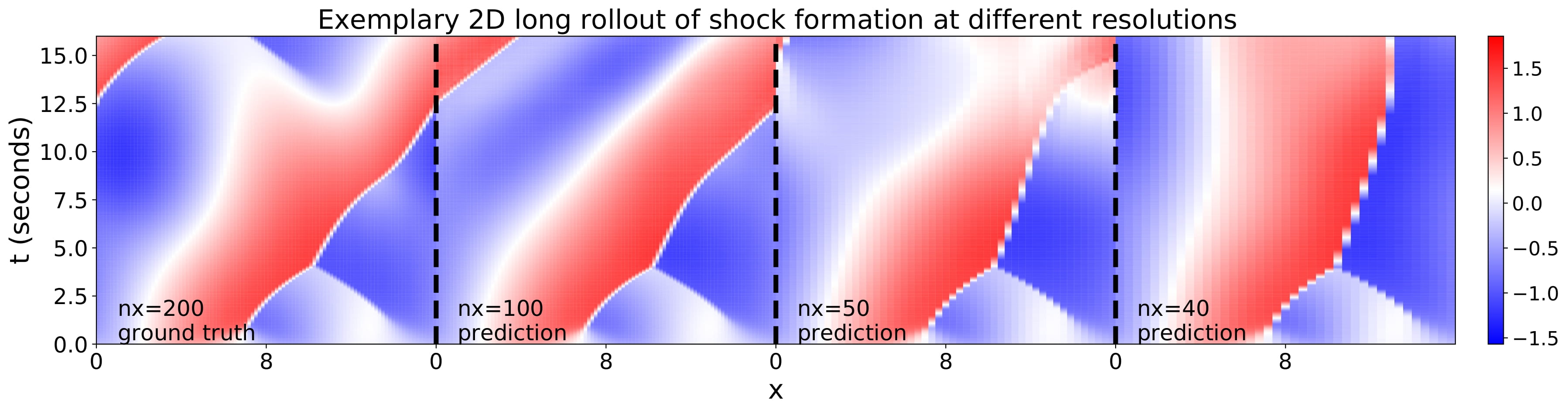}
    \caption{Demonstration of shock formation using MP-PDE from different training data resolutions. Credits: Brandstetter et al.~\citep{brandstetterMessagePassingNeural2022}.}
\end{figure}

The same data is displayed in 2D to show the time evolution. After about 7.5s, the error accumulation is large enough to visibly diverge from the ground truth. The predictions become unreliable due to error accumulation.

In practice, this survival time should be empirically found (as seen here) to determine how long the solution is reliable. However, the ground truth would be needed for comparison, rendering this as another chicken-egg problem.

\begin{table}[htb]
    \centering
    \begin{tabular}{|c|c|*{6}{|c}|}
    \hline
    \multicolumn{2}{|c|}{} & \multicolumn{4}{c|}{Accumulated Error} & \multicolumn{2}{c|}{Runtime [s]} \\
    \hline
    \multicolumn{2}{|c|}{$(n_{t},n_{x})$} & WENO5 & FNO-RNN & FNO-PF & MP-PDE & WENO5 & MP-PDE \\
    \hline
    \textbf{E1} & (250,100) & 2.02 & 11.93 & 0.54 & 1.55 & 1.9 & 0.09 \\
    \textbf{E1} & (250, 50) & 6.23 & 29.98 & 0.51 & 1.67 & 1.8 & 0.08 \\
    \textbf{E1} & (250, 40) & 9.63 & 10.44 & 0.57 & 1.47 & 1.7 & 0.08 \\
    \textbf{E2} & (250,100) & 1.19 & 17.09 & 2.53 & 1.58 & 1.9 & 0.09 \\
    \textbf{E2} & (250, 50) & 5.35 & 3.57 & 2.27 & 1.63 & 1.8 & 0.09 \\
    \textbf{E2} & (250, 40) & 8.05 & 3.26 & 2.38 & 1.45 & 1.7 & 0.08 \\
    \textbf{E3} & (250,100) & 4.71 & 10.16 & 5.69 & 4.26 & 4.8 & 0.09 \\
    \textbf{E3} & (250, 50) & 11.71 & 14.49 & 5.39 & 3.74 & 4.5 & 0.09 \\
    \textbf{E3} & (250, 40) & 15.97 & 20.90 & 5.98 & 3.70 & 4.4 & 0.09 \\
    \hline
    \end{tabular}
    \caption{Table of experiment results adapted from paper. Credits:~\citet{brandstetterMessagePassingNeural2022}.}
\end{table}

\paragraph{Abbreviations}

\begin{table}[htb]
    \centering
    \begin{tabular}{|c|l|}
    \hline
    Shorthand & Meaning \\
    \hline
    \textbf{E1} & Burgers' equation without diffusion \\
    \textbf{E2} & Burgers' equation with variable diffusion \\
    \textbf{E3} & Mixed equation, see below \\
    \(n_{t}\) & Temporal resolution \\
    \(n_{x}\) & Spatial resolution \\
    WENO5 & Weighted Essentially Non-Oscillatory (5th order) \\
    FNO-RNN & Recurrent variation of FNO from original paper \\
    FNO-PF & FNO with the pushforward trick added \\
    MP-PDE & Message passing neural PDE solver \\
    \hline
    \end{tabular}
\end{table}

The authors form a general PDE in the form:
\begin{equation}
    \partial_{t}u + \partial_{x}(\alpha u^{2} - \beta \partial_{x} u + \gamma \partial_{xx} u)(t,x) = \delta (t,x)
\end{equation}
with initial condition:
\begin{equation}
    u(0,x) = \delta(0,x)
\end{equation}
such that \(\theta_{\text{PDE}} = (\alpha, \beta, \gamma)\) and different combinations of these result in the heat equation, Burgers' equation, and the KdV equation. \(\delta\) is a forcing term, allowing for greater variation in the equations being tested.

For this same experiment, the error and runtimes were recorded when solving using WENO5, the recurrent variant of the FNO (FNO-RNN), the FNO with the pushforward trick (FNO-PF), and the MP-PDE.

\emph{The pushforward trick is successful in mitigating error accumulation.}

Comparing the accumulated errors of FNO-RNN and FNO-PF across all experiments highlights the advantage of the pushforward trick. While the MP-PDE outperforms all other tested methods in the two generalization experiments \textbf{E2} and \textbf{E3}, the FNO-PF is most accurate for \textbf{E1}.

When solving a single equation, the FNO likely performs better, though both FNO-PF and MP-PDE methods outperform WENO5.

\emph{Neural solvers are resolution-invariant.}

As $$n_{x}$$ is decreased, WENO5 performs increasingly worse whereas all the neural solvers remain relatively stable.

\emph{Neural solver runtimes are constant to resolution.}
Additionally, the runtimes of WENO5 decrease (likely proportionally) since fewer steps require fewer calculations, but the MP-PDE runtimes again appear relatively stable.

\subsection{Comparing Interpretations}
The way the MP-PDE is constructed parallels how both GRAND and the PDE-GCN are built. All three architectures follow a basic premise of mirroring the MOL and describe certain mechanisms in their respective systems which mimic spatial discretisations and temporal discretisations.

The spatial derivative is discretized by a GNN in the MP-PDE and by the message passing algorithm (consisting of node and edge updates within one layer of a GNN) in the GRAND and PDE-GCN. In the MP-PDE, the spatial derivatives are in effect parameterized by the node and edge updates (the former which Brandstetter et al. highlight takes the difference in solutions $u_{i}=u_{j}$) detailed above, both of which are generic MLPs. In comparison, both GRAND and PDE-GCN (using the diffusion variant) come to comparable formulas when discretising using the forward Euler method.

The GRAND paper derives the following, where $\tau$ is a temporal step, $\mathbf{x}$ is the diffusion equation, and $\mathbf{A}$ is the attention matrix~\citep{chamberlainGRANDGraphNeural2021a}:

\begin{equation}
\mathbf{x}^{(k+1)}=(\mathbf{I} + \tau \mathbf{A}(\mathbf{x}^{(k)}))\mathbf{x}^{(k)}
\end{equation}

which, when modified, results in:

\begin{equation}
\mathbf{x}^{(k+1)}=\mathbf{x}^{(k)} + \tau \mathbf{x}^{(k)} \mathbf{A}(\mathbf{x}^{(k)})
\end{equation}

The PDE-GCN defines manifold operators discretized onto graphs. The update is defined as the following, where $\mathbf{G}$ is the gradient operator, $\mathbf{K}$ is a $1 \times 1$ trainable convolution kernel, $\sigma$ is the activation function, $\tau$ is the temporal step, and $\mathbf{x}$ is the diffusion equation~\citep{eliasofPdegcnNovelArchitectures2021}:

\begin{equation}
\mathbf{x}^{(k+1)}=\mathbf{x}^{(k)}-\tau \mathbf{G}^{T} \mathbf{K}^{T}_{k} \sigma (\mathbf{K}_{k} \mathbf{G} \mathbf{x}^{(k)})
\end{equation}

The structure of these latter two models shares many similarities, though where GRAND naturally results in a graph attention network, the PDE-GCN results in a graph convolutional network.

The temporal update for the MP-PDE relies on the 1D CNN outputting a temporal bundle, whereas GRAND and PDE-GCN regard their respective layer indexes to be the discretised time steps.

These are examples of how spatial and temporal discretisations can result in unique architectures. The PDE-GCN outperforms GRAND on at least two out of three of the popular Cora, SiteSeer, and PubMed benchmarks. However, the MP-PDE has a different objective altogether; while the PDE-GCN and GRAND output a single graph result (which is fed through a convolutional layer for node classification tasks), the MP-PDE iteratively produces results through time. This iterative requirement also requires that the temporal update must be retrievable and therefore must diverge from Ruthotto et al.'s original interpretation of time steps as layers adopted by the other two models. The MP-PDE instead appears to rely on the neural networks in both node and edge updates to learn spatial derivatives over multiple layers. An interesting experiment would be to apply the other two techniques to the same testing data as PDE-GCN and compare accuracies at a specific point in time (see future directions).

\section{Conclusion}

\paragraph{Future directions}
The authors conclude by discussing some future directions.

For example, the MP-PDE can be modified for PDE retrieval (which they call parameter optimization). There is some precedence for this: Cranmer et al. develop a method which fits a symbolic regression model (e.g., PySR, eureqa) to the learned internal functions of a GNN~\citep{cranmerDiscoveringSymbolicModels2020}. Alternatively, the MP-PDE's capacity for generalization means that biasing the model with a prior to determine coefficients could be as simple as training on an example instance of the predicted equation, fitting this model on real-world data (much like a fine-tuning process), and extracting the $\theta_{PDE}$ parameters.

The one-step loss, which is the basis of the adversarial-style loss, is also used in reinforcement learning, which frequently uses deep autoregressive models. Other formulations which borrow from reinforcement learning (where distribution shifts are quite common) and other fields could prove successful as well. Transformer-based natural language processing is now capable of capturing extremely long sequence dependencies and generating coherent long-form text. Since \href{https://graphdeeplearning.github.io/post/transformers-are-gnns/}{Transformers are GNNs} which use attention to aggregate neighborhoods, this may be a viable avenue to explore.

Adaptive time-stepping is another avenue which could make the model more efficient and accurate by taking large steps over stable/predictable solution regions and smaller steps over changing/unpredictable solution regions. The choice of a CNN for the decoder works well over regular inputs and outputs, but other options like attention-based architectures could potentially weigh the outputted node embeddings such that the model might learn different time steps. Some care would have to be taken with temporal bundling in this case, since the resulting vectors would be potentially irregular in time.

In addition, while the GRAND architecture is designed for a single output, adapting it to suit an iterative solver may prove fruitful since the attention mechanism would encode spatial awareness. The motivation for this choice is that a sparse attention matrix might be able to provide a more global solution.

\paragraph{Ongoing Challenges}

While there are numerous diverse branches of development, key challenges remain:

\begin{itemize}
    \item Unified and appropriate evaluation metrics
    \begin{itemize}
        \item Currently, mean squared error (or root mean squared error) is implemented as the choice of loss in not only MP-PDE, but most named networks herein. However, it is unclear whether this is the best measure of correctness to solving a PDE since the specific values of the solution evaluated at the discretised points will depend on the discretisation method. An interesting further study would be to use the MP-PDE and test it on data generated from multiple numerical solvers. Additionally, Brandstetter et al. identify a metric called survival time which defines the length of time before the predicted solution diverges past a specified error threshold. Such metrics are important from a user's perspective when choosing between architectures, but there has yet to be a unified set of metrics in literature and so we lack convenient benchmarking.
    \end{itemize}
    
    \item Understanding choices in network architecture
    \begin{itemize}
        \item Given an end goal of using neural PDE solvers in practical settings, a major barrier for not only MP-PDE but for GRAND and PDE-GCN as well are the difficulties in choosing network parameters. While the proposed MP-PDE sheds light on certain choices like the message passing function and encoder-processor-decoder structure, it does not address some pragmatic decisions. For example, the 6 message passing layers in the MP-PDE appears relatively arbitrary which is a complaint shared in many machine learning methods. Because of the resulting upfront work in optimising the chosen model to determine what works for a new problem setting, the time cost of implementing it can be prohibitively high in comparison to the relative convenience of the many numerical solvers. One avenue of research to address this concern is neural architecture searching, where the design of neural architectures is discovered rather than manually specified. However, there is still a long way to go as many automated searches require significant compute to test the parameter space adequately.
    \end{itemize}
    
    \item The chicken and the egg
    \begin{itemize}
        \item As impressive as many novel neural methods may be, it remains that training data comes from classical methods. One of the largest open questions (which also drives the need for generalisation) is how we can design neural solvers which require as little data as possible. Transfer learning, curriculum learning, and techniques to encourage generalisation (as seen with the MP-PDE) are all steps toward addressing this problem, but no significant success has been seen from any one in particular.
    \end{itemize}
\end{itemize}

\paragraph{Remarks}
In their paper “Message Passing Neural PDE Solver”, Brandstetter at al. present a well-motivated neural solver based on the principle of message passing. The key contributions are the end-to-end network capable of one-shot generalization, and the mitigation of error accumulation in autoregressive models via temporal bundling and the pushforward trick. Note that the latter are self-contained can be applied to other architectures (as in the FNO-PF), providing a valuable tool to improve autoregressive models.

\bibliography{bibliography}
\bibliographystyle{iclr2022_conference}
\end{document}

%% file: math_commands.tex
\usepackage{amsmath,amsfonts,bm}

\def\eqref#1{equation~\ref{#1}}
\def\1{\bm{1}}

\DeclareMathAlphabet{\mathsfit}{\encodingdefault}{\sfdefault}{m}{sl}
\SetMathAlphabet{\mathsfit}{bold}{\encodingdefault}{\sfdefault}{bx}{n}